\documentclass[conference,letterpaper]{IEEEtran}
\IEEEoverridecommandlockouts

\usepackage{graphicx}
\usepackage{booktabs}
\usepackage{xcolor}
\usepackage{amsmath}
\usepackage{cite}
\usepackage[hidelinks]{hyperref}
\usepackage{xurl}
\usepackage[T1]{fontenc}
\usepackage[utf8]{inputenc}
\usepackage{textcomp}
\usepackage{mathptmx}
\usepackage{CJKutf8}
\newcommand{\zh}[1]{\begin{CJK}{UTF8}{gbsn}#1\end{CJK}}

\makeatletter
\renewcommand{\footnoterule}{\kern-3pt\hrule width 0.4\columnwidth height 0.4pt\kern 2.6pt}
\makeatother

\begin{document}

\title{Lao Siji\textsuperscript{*}\thanks{\textsuperscript{*}\emph{Lao Siji} (\zh{老司机}): Chinese slang for ``an experienced driver''.}: Adapting openpilot to Closed-Loop End-to-End Driving in Neural Reconstruction}

\author{
  \IEEEauthorblockN{Chengzhi Gao, Yongzhi Liu, Jinchang Xu, Weichao Zhuang$^{\dagger}$, Guodong Yin}
  \IEEEauthorblockA{School of Mechanical Engineering, Southeast University, Nanjing, China \\
  \texttt{Gaochengzhi1999@gmail.com}, \texttt{wezhuang@seu.edu.cn} \\
  $^{\dagger}$Corresponding author}
}

\maketitle

\begin{abstract}
We present Lao Siji, an end-to-end driving policy that ranks third in the 2026 NVIDIA AlpaSim E2E Driving Challenge, ahead of most entries built on foundation models. Instead of training from scratch on a general-purpose vision backbone, we adapt openpilot, an open-source L2 production model. We freeze its vision encoder and fine-tune the policy layers with an ego-state adapter and a route memory. On closed-loop rollouts in neural-reconstructed scenes, we supervise the policy with a privileged map teacher for lateral self-correction and a body-extent loss for road and obstacle clearance, while distilling the base model's remaining outputs to retain car following. Ablations show that self-correction samples carry the gain, while the clearance loss substitutes for them where no teacher exists. Longitudinal safety instead rests on the base model's lead-vehicle estimate behind a rule-based speed layer. Longer training, imitating human logs, and human-log speed targets yield either no measurable gain or adverse effects. Without further training, Lao Siji also places among the leaders on HUGSIM and Odyssey. The final training takes only one GPU-hour and shifts the weights by 3.9\% from the shipped model. Code will be released after review.
\end{abstract}


\section{Introduction}
\label{sec:intro}

\begin{figure*}[t]
\centering
\includegraphics[width=0.9\textwidth]{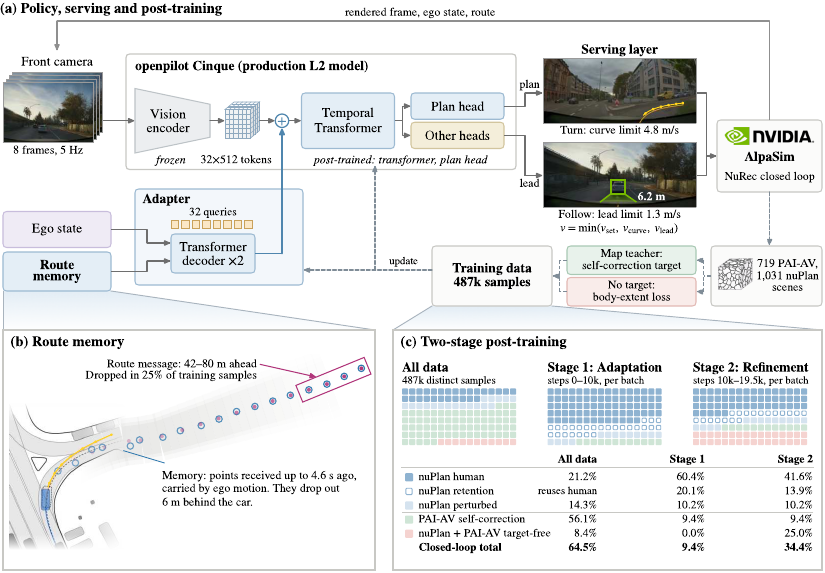}
\caption{Overview of Lao Siji. (a) Post-training adapter and policy layers with a frozen openpilot vision encoder; states reached in AlpaSim rollouts return to the training data (dashed), with a self-correction target from a map teacher or with no target and the body-extent loss alone. (b) Route memory extends the route message to the ego vehicle (offsets exaggerated for illustration). (c) Two-stage post-training: composition of the collected samples and of a batch of 128 samples; Stage 2 adds the target-free samples.}
\label{fig:system}
\end{figure*}

Since UniAD~\cite{uniad} established the paradigm of jointly optimizing perception, prediction, and planning in an end-to-end framework, the field has expanded along several major directions, including trajectory-scoring planners~\cite{hydramdp,drivor}, vision-language-action models~\cite{autovla,alpamayo_r1}, and world-model-based planners~\cite{epona,wa_jepa}. Across these lines, most recent approaches pair a general-purpose vision backbone with a driving head and train over weeks on multi-GPU clusters~\cite{epona,vavam}, imposing a steep compute barrier for resource-constrained groups.

Alongside this compute cost, research has centered on model architecture while leaving data curation and post-training far less explored~\cite{codevilla2019,hiddenbiases_data,lead}. Although complex designs have pushed scores on open-loop benchmarks such as NAVSIM~\cite{navsim} near human performance~\cite{drivor}, these scores reflect closed-loop driving capability only weakly (Table~\ref{tab:main}). Photorealistic closed-loop simulators built on neural scene reconstruction, such as HUGSIM~\cite{hugsim}, AlpaSim~\cite{alpasim}, and Odyssey~\cite{odyssey}, help close this gap.

We study closed-loop post-training in the 2026 NVIDIA AlpaSim End-to-End Driving Challenge~\cite{alpasim_challenge}.
Our solution builds on openpilot~\cite{openpilot}, an open-source L2 driver-assistance model with a front-camera encoder pre-trained on large-scale real-world driving data.
The model keeps its lane and follows a lead vehicle but cannot turn at intersections; before closed-loop training, lateral failures outnumber front collisions by 2.5 to 3.8 times (Table~\ref{tab:cause}).
The closed-loop supervision therefore targets the lateral path only.
The vision encoder is frozen, and 5\,\% of the parameters are trained on rollouts in neural-reconstructed scenes.
Following dataset aggregation (DAgger)~\cite{dagger}, a privileged map teacher corrects the lateral path on the states the policy reaches, and a body-extent loss enforces road and obstacle clearance where the teacher gives no target.
Speed remains with the lead-vehicle estimate of the shipped model behind a rule-based layer.

Ablations attribute the gain to the lateral self-correction samples; without them, the policy falls back close to the level of offline post-training.
The clearance loss improves a policy trained without a teacher but adds no measurable gain once teacher targets are present.
Longer training and human-log targets yield no resolved gain.
The resulting policy, \emph{Lao Siji}, ranks third in the challenge, ahead of most entries built on foundation models, with a final training stage of about one GPU-hour on a single NVIDIA RTX 6000D.
Without further training, it is among the leading methods on HUGSIM~\cite{hugsim} and Odyssey~\cite{odyssey} and raises the NAVSIM~\cite{navsim} EPDMS of the shipped model from 80.5 to 88.7.

Our contributions are as follows:
\begin{enumerate}
  \item We present a compute-efficient post-training recipe that adapts a production L2 model into a competitive end-to-end driving policy with strong cross-benchmark transfer, without training a vision stack from scratch (Sec.~\ref{sec:method}).
  \item We conduct a systematic ablation across closed-loop states, supervisory signals, and the serving layer, isolating each component's strengths, redundancies, and boundaries between learned planning and inherited control (Sec.~\ref{sec:main} and~\ref{sec:ablation}).
  \item We document negative results across data selection, rollout scaling, and training duration to inform post-training design under compute constraints (Sec.~\ref{sec:nothelp}).
\end{enumerate}

\section{Related Work}
\label{sec:related}

Most post-training methods for end-to-end driving operate on logged states: Hydra-MDP~\cite{hydramdp} and DrivoR~\cite{drivor} distill rule-based trajectory scores into the planner, and AutoVLA~\cite{autovla} and Alpamayo-R1~\cite{alpamayo_r1} use trajectory scores as rewards for reinforcement fine-tuning.
Training on the states reached by the policy itself follows two paradigms.
Online reinforcement learning runs on privileged state inputs (GigaFlow~\cite{gigaflow}, CaRL~\cite{carl}) or on camera input in reconstructed scenes (RAD~\cite{rad}, AlpaGym~\cite{alpagym}) and requires large rollout budgets.
Dataset aggregation (DAgger)~\cite{dagger} and privileged distillation~\cite{lbc,roach} let an expert teacher label the states visited by the student.

Three recent methods are closest to ours.
RoaD~\cite{road} guides the rollouts of the policy toward the logged human trajectory, OPTED~\cite{opted} labels the states of a frozen-backbone student with a reinforcement-learning teacher in AlpaSim, and DriveZero~\cite{drivezero} distills such a teacher into a camera policy on logged frames, but none of them had released runnable code and weights at the time of submission.
All three supervise a complete trajectory, speed included, and report the pipeline as a whole; we supervise the lateral path, give no deceleration target, and ablate the components of the recipe one at a time.

\section{Method}
\label{sec:method}

Lao Siji extends a shipped driving model with navigation inputs and post-trains it in reconstructed scenes by dataset aggregation (DAgger)~\cite{dagger} (Fig.~\ref{fig:system}).
The longitudinal control of the shipped model is in production use on the road, and lateral failures are the main source of zero-score scenes before closed-loop training (Table~\ref{tab:cause}).
We therefore supervise only the lateral path and the clearance of the vehicle on the states the policy reaches (Sec.~\ref{sec:priv}) and retain the skills of the shipped model with dedicated samples (Sec.~\ref{sec:mix}).

\subsection{Base Model and Adapter}
\label{sec:frozen}\label{sec:route}
We build upon openpilot Cinque~\cite{openpilot}, a production L2 driving model (Fig.~\ref{fig:system}(a)).
It consists of a frozen vision encoder (352.9M parameters), a post-trained temporal transformer with a plan head (16.2M), and auxiliary frozen output heads (12.5M, including lead-vehicle estimation for adaptive cruise control).
Since the base model accepts no route, we introduce an adapter that injects the ego state (velocity, acceleration, past four poses, and turn command) and navigation route into the vision tokens.

The input is eight frame pairs 0.2\,s apart, and the encoder emits 32 tokens of width 512 per pair.
The adapter comprises two transformer-decoder branches of identical design: 32 learned queries attend to branch input tokens via a two-layer decoder, and a linear projection yields a $32{\times}512$ bias that is added to these tokens one to one.
The ego branch (2.5M parameters, width 256) processes four tokens embedded from the ego state via an MLP.
The route branch (0.6M parameters, width 128) processes an ego token and up to 40 route waypoints, each embedded with learned positional encodings.
Both linear output layers are zero-initialized, ensuring post-training begins strictly from the base model's original distribution.
Waypoints are drawn from a parameter-free route memory that accumulates incoming messages (each covering about 42 to 80\,m ahead), tracks ego displacement, and provides points from 6\,m behind to 72\,m ahead of the car (Fig.~\ref{fig:system}(b)); route inputs are dropped for 25\,\% of training samples for robustness.
The policy outputs an 8-pose 4\,s plan and a lead-vehicle estimate, which the serving layer (Sec.~\ref{sec:serving}) converts into a tracking reference.

\begin{figure}[t]
\centering
\includegraphics[width=\linewidth]{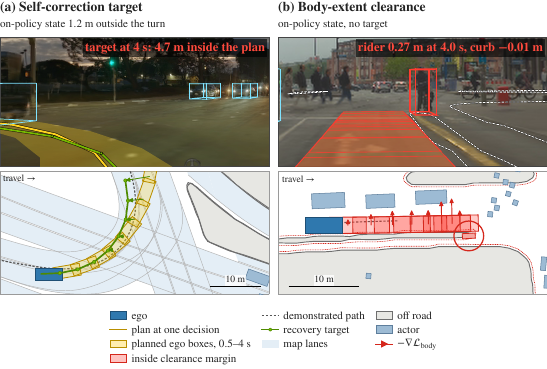}
\caption{Supervision on self-reached states. (a) Self-correction target: a recovery trajectory from a state 1.2\,m outside a turn. (b) Body-extent clearance: on a target-free state, the clearance loss backpropagates normal to the path, moving the plan away from the curb and an actor.}
\label{fig:body}
\end{figure}

\subsection{Closed-Loop Rollouts and Supervision}
\label{sec:rollouts}\label{sec:priv}
\textbf{Iterative rollouts.}
In each round, the policy is trained from the shipped weights on all samples accumulated so far and rolled out once across the training scenes; only states before the first infraction, as the scorer of the benchmark defines it (Sec.~\ref{sec:setup}), are kept (in nuPlan scenes, states within 4\,m of the logged path). Earlier policies supply states, and only the final policy is served.
The initial policy is trained on human logs alone (Sec.~\ref{sec:mix}).
The released recipe runs five sequential rounds: the first in 1{,}031 scenes of the AlpaSim nuPlan track, followed by four rounds in 719 PAI-AV scenes~\cite{pai_av} disjoint from evaluation scenes.
The states of a round are supervised in one of two modes (Fig.~\ref{fig:body}): with a self-correction target, or with no target and a clearance loss alone.

\textbf{Self-correction targets.}
Self-correction samples receive a trajectory target from a privileged map teacher (Fig.~\ref{fig:body}(a)).
Its reference is the route centerline shifted by the lateral offset of the human driver (smoothed over 10\,m, at most 0.5\,m).
From the state of the policy, a Frenet quintic closes the lateral and heading errors over the shortest horizon in $\{2.0, 2.5, 3.0\}$\,s whose lateral acceleration stays within 4\,m/s$^2$.
A state receives no target if no horizon meets this bound, if it is more than 5\,m or 60\textdegree{} off the reference or slower than 2\,m/s, or if the path passes an obstacle closer than the human did or comes within 1\,m + 0.84\,s $\times$ speed of a lead vehicle where the served plan does not.
Along the path, the target keeps the arc lengths of the served plan and moves them only forward: where the policy lags with nothing to yield to, it restores the speed of the human or of same-direction traffic, since the plan of the shipped model decays toward a stop in closed loop (Table~\ref{tab:main}) and the benchmark penalizes slow driving.
No target slows the policy down; decoupling the lateral path from the speed has precedents~\cite{tfpp,simlingo} and suits log-replay evaluation.

\textbf{Body-extent clearance.}
Target-free samples receive no trajectory target and are supervised only by penalties on the plan (Fig.~\ref{fig:body}(b)).
Along the 4\,s plan, the vehicle footprint must keep 0.3\,m from road edges and from simulator-derived agent boxes in its path; in the nuPlan round, it is also penalized for entering oncoming lanes and lanes off the route.
These penalties backpropagate solely through the normal component of each planned pose: they can steer or bend the trajectory away from obstacles, but cannot shorten it, so the policy cannot reduce the loss by decelerating.
Self-correction samples carry the same clearance penalties but no lane penalties: their teacher keeps the human's lateral placement, which occasionally crosses a lane line, and penalizing lanes as well would leave the policy too rigid.
The released recipe designates whole rounds as either target-free (nuPlan and third PAI-AV round, 41k samples total) or self-correction (remaining three PAI-AV rounds, 273k samples total).

\begin{table}[t]
\centering
\caption{Zero-score scenes by cause on the 433-scene AlpaSim exam.}
\label{tab:cause}
\footnotesize
\setlength{\tabcolsep}{5pt}
\begin{tabular}{@{}lrrrr@{}}
\toprule
Model & Lateral & Longitudinal & Other & Zeros \\
\midrule
\multicolumn{5}{@{}l}{\emph{Not trained in closed loop}} \\
UniAD~\cite{uniad} & 236\,{\scriptsize(68\%)} & 65\,{\scriptsize(19\%)} & 48\,{\scriptsize(14\%)} & 349 \\
AutoVLA~\cite{autovla} & 189\,{\scriptsize(63\%)} & 50\,{\scriptsize(17\%)} & 60\,{\scriptsize(20\%)} & 299 \\
DrivoR~\cite{drivor} & 140\,{\scriptsize(65\%)} & 55\,{\scriptsize(25\%)} & 22\,{\scriptsize(10\%)} & 217 \\
WA-JEPA~\cite{wa_jepa} & 127\,{\scriptsize(52\%)} & 36\,{\scriptsize(15\%)} & 82\,{\scriptsize(33\%)} & 245 \\
openpilot~\cite{openpilot} & 132\,{\scriptsize(53\%)} & 38\,{\scriptsize(15\%)} & 77\,{\scriptsize(31\%)} & 247 \\
\quad after offline post-training & 56\,{\scriptsize(62\%)} & 15\,{\scriptsize(16\%)} & 20\,{\scriptsize(22\%)} & 91 \\
\midrule
\multicolumn{5}{@{}l}{\emph{Post-trained in closed loop}} \\
py123d\_garage~\cite{garage} & 29\,{\scriptsize(46\%)} & 31\,{\scriptsize(49\%)} & 3\,{\scriptsize(5\%)} & 63 \\
Lao Siji (ours) & 16\,{\scriptsize(46\%)} & 15\,{\scriptsize(42\%)} & 4\,{\scriptsize(12\%)} & 35 \\
\quad w/o lead limit & 19\,{\scriptsize(18\%)} & 76\,{\scriptsize(74\%)} & 8\,{\scriptsize(7\%)} & 103 \\
\bottomrule
\end{tabular}
\\[2pt]
\begin{minipage}{\linewidth}
\scriptsize Parentheses: share of row zeros. Lateral: off-road or route exit; Longitudinal: at-fault front collision; Other: side contacts and unclassified ends. openpilot runs through our serving layer; offline post-training uses human logs only; Lao Siji: three-seed means, rounded; shares from the unrounded means.
\end{minipage}
\end{table}

\subsection{Training Samples and Schedule}
\label{sec:mix}
Table~\ref{tab:samples} lists the training samples, and Fig.~\ref{fig:system}(c) their shares.
Besides the two closed-loop kinds, three kinds come from human driving logs of nuPlan~\cite{nuplan}.
\emph{Human} samples are the 103k frames of the NAVSIM navtrain split~\cite{navsim} with the logged future as the target; they teach the policy to use the route and the ego state.
\emph{Retention} samples are 25\,\% of the human frames in each batch, fed with both new inputs switched off and with the plan of the shipped model as the target; they are meant to keep the network without the adapter close to the shipped one (the route drop of Sec.~\ref{sec:frozen} is separate: it removes the route alone, on the other samples).
\emph{Perturbed} samples are 70k of the human frames seen from a perturbed pose (heading errors of a few degrees, lateral offsets of up to 1\,m), with the logged future as the target.

\newsavebox{\samplesbox}
\begin{table}[t]
\centering
\caption{Training samples and their supervision.}
\label{tab:samples}
\footnotesize
\setlength{\tabcolsep}{2.4pt}
\sbox{\samplesbox}{%
\begin{tabular}{@{}llll@{}}
\toprule
Sample & States & Target & $\mathcal{L}_{\mathrm{body}}$ against \\
\midrule
Human & human log & logged future & drivable area \\
Retention & log, new inputs off & shipped plan & none \\
Perturbed & log, perturbed pose & logged future & drivable area \\
Self-correction & own rollouts & teacher path & road edge, agents \\
Target-free & own rollouts & none & road edge, agents, lanes$^{*}$ \\
\bottomrule
\end{tabular}}%
\usebox{\samplesbox}\\[2pt]
\begin{minipage}{\wd\samplesbox}\scriptsize $^{*}$Lanes: nuPlan round only (Sec.~\ref{sec:priv}).\end{minipage}
\end{table}

\textbf{Schedule.}
Every training starts from the shipped weights and runs on batches of 128 samples.
The first policy of the chain sees only the log samples, for 10k steps.
Every later training runs for 19.5k steps in two stages (Fig.~\ref{fig:system}(c)).
In stage~1 (adaptation, 10k steps), every sample that carries a target is drawn: the policy learns to follow the route and to correct itself on its own states.
In stage~2 (refinement, 9.5k steps), the target-free samples enter as a quarter of each batch, and the learning rates of the network and of the ego branch of the adapter drop tenfold.
The target-free samples penalize a plan but do not say where to drive, so they enter once the policy follows the route, at a rate that keeps it there.

\textbf{Loss.}
Both stages minimize
\begin{equation}
\mathcal{L}=\mathcal{L}_{\mathrm{imit}}+\lambda_{\mathrm{c}}\mathcal{L}_{\mathrm{cons}}+\lambda_{\mathrm{d}}\mathcal{L}_{\mathrm{dist}}+\lambda_{\mathrm{b}}\mathcal{L}_{\mathrm{body}}.
\label{eq:loss}
\end{equation}
$\mathcal{L}_{\mathrm{imit}}$ is a Huber loss (threshold 1 on the scaled error) on the position and heading errors of the eight plan poses against the target, normalized by a tolerance that grows with the horizon, on the samples that carry one.
$\mathcal{L}_{\mathrm{body}}$ is the hinge $\max(0,\,0.3\,\mathrm{m}-d)$ on the clearance $d$ of the planned footprint, as in Table~\ref{tab:samples} (in the nuPlan round, together with the lane penalties of Sec.~\ref{sec:priv}), with $\lambda_{\mathrm{b}}=30$ on closed-loop samples and $10$ on log samples.
$\mathcal{L}_{\mathrm{cons}}$ ($\lambda_{\mathrm{c}}=3$) is the same pose error against the plan of the shipped model on the retention samples.
$\mathcal{L}_{\mathrm{dist}}$ ($\lambda_{\mathrm{d}}=30$) distills all outputs other than the plan to those of the shipped model on every sample, the target-free ones included: the weights of these heads are frozen, but they read the post-trained temporal features, and the serving layer relies on the lead-vehicle estimate.

We train with AdamW at learning rates of $3{\times}10^{-5}$ for the network and $3{\times}10^{-4}$ for both branches of the adapter, with a cosine decay.
The final training moves the trained weights by 3.9\,\% from the shipped ones (relative $L_2$ distance).

\subsection{Serving Layer}
\label{sec:serving}
At every 0.1\,s decision, the serving layer converts the plan and the lead estimate of the policy into the reference trajectory tracked by the controller of the simulator.
The reference keeps the path $\gamma(s)$ of the plan, and the layer sets the timing along it:
\begin{align}
p_k &= \gamma\bigl(s(t_k)\bigr), \qquad s(t)=\int_0^{t} v(\tau)\,d\tau, \label{eq:serve}\\
v^{*} &= \min\bigl(v_{\mathrm{set}},\, v_{\mathrm{curve}},\, v_{\mathrm{lead}}\bigr), \label{eq:vstar}
\end{align}
where the speed $v$ approaches the target $v^{*}$ from the ego speed within braking and jerk bounds, $v_{\mathrm{set}}$ is the greater of the ego speed at the handover and the planned speed 2 to 4\,s ahead, $v_{\mathrm{curve}}=\sqrt{a_{\mathrm{lat}}/\kappa}$ limits the lateral acceleration on the curvature $\kappa$ ahead, and $v_{\mathrm{lead}}$ is the lead limit: the lead-vehicle controller of openpilot on the lead estimate of the policy.
The plan itself is not passed on, because the controller tracks positions without a penalty on steering: asked for more deceleration than the brakes give, it steers and spins the vehicle (32 of the 433 exam scenes with the raw plan, Sec.~\ref{sec:ablation}).
Where the path turns too sharply for the served speed, the reference falls back to a path-tracking law with bounded curvature.
Two rules are specific to the challenge: at a highway-speed handover with the route in an adjacent lane, the route is held in the ego lane, and through a U-turn the served path follows the route.
Removing them is not resolved on the exam (Sec.~\ref{sec:ablation}), and the U-turn rule is off on all other benchmarks.
PAI-AV rollout rounds and evaluations share this layer so that collected states match the served policy, whereas the 2\,Hz nuPlan round passes the plan to the controller unchanged.

\section{Experiments}
\label{sec:exp}

\begin{table*}[t]
\centering
\caption{Comparison with baselines on closed-loop benchmarks and NAVSIM.}
\label{tab:main}
\footnotesize
\setlength{\tabcolsep}{3.6pt}\setlength{\fboxsep}{1.5pt}\setlength{\fboxrule}{0.4pt}
\begin{tabular}{@{}l rrr rrrrr r rr@{}}
\toprule
Model & \multicolumn{3}{c}{AlpaSim PAI-AV} & \multicolumn{5}{c}{HUGSIM} & \multicolumn{1}{c}{Odyssey} & \multicolumn{2}{c}{NAVSIM} \\
\cmidrule(lr){2-4}\cmidrule(lr){5-9}\cmidrule(lr){10-10}\cmidrule(lr){11-12}
 & Score\,$\uparrow$ & Zeros\,$\downarrow$ & At-fault\,$\downarrow$ & NC\,$\uparrow$ & DAC\,$\uparrow$ & TTC\,$\uparrow$ & Rc\,$\uparrow$ & HD-Score\,$\uparrow$ & RouteDS\,$\uparrow$ & PDMS\,$\uparrow$ & EPDMS\,$\uparrow$ \\
\midrule
UniAD~\cite{uniad} & 0.119 & 349 & 238 & 0.629 & 0.923 & 0.473 & 0.493 & 0.327 & 4.6 & -- & -- \\
AutoVLA~\cite{autovla} & 0.185 & 299 & 196 & 0.630 & 0.952 & 0.542 & 0.392 & 0.280 & 13.6 & 89.5 & 89.2 \\
DrivoR~\cite{drivor} & 0.349 & 217 & 142 & 0.584 & 0.954 & 0.503 & 0.455 & 0.322 & 24.4 & \textbf{93.7} & \fbox{90.1} \\
WA-JEPA~\cite{wa_jepa} & 0.331 & 245 & 124 & \fbox{0.779} & \fbox{0.961} & 0.718 & \textbf{0.539} & \textbf{0.461} & 36.2 & \fbox{91.7} & \textbf{91.7} \\
\midrule
py123d\_garage$^{\star\dagger}$~\cite{garage} & \fbox{0.774} & \fbox{63} & \fbox{60} & \textbf{0.793} & \textbf{0.962} & \textbf{0.743} & 0.522 & 0.457 & \textbf{55.2} & 88.6 & 88.6 \\
openpilot~\cite{openpilot} & 0.377 & 247 & 127 & 0.819 & 0.880 & 0.792 & 0.217 & 0.182 & 11.2 & 82.0 & 80.5 \\
Lao Siji$^{\dagger\ddagger}$ (ours) & \textbf{0.907} & \textbf{35.3} & \textbf{29.7} & 0.771 & 0.927 & \fbox{0.724} & \fbox{0.535} & \fbox{0.460} & \fbox{49.4} & 88.5 & 88.7 \\
\bottomrule
\end{tabular}
\\[2pt]
\begin{minipage}{\textwidth}
\scriptsize NC: no collision; DAC: drivable-area compliance; TTC: time to collision; Rc: route completion. PDMS, EPDMS: the driving scores of NAVSIM~\cite{navsim} and of its second version~\cite{pseudosim}, on navtest. Best in bold, second best boxed; Lao Siji: three-seed means. $^{\dagger}$Trained on PAI-AV scenes. $^{\ddagger}$Training logs shared with 68 of the 100 Odyssey routes. $^{\star}$Fed target points from the recorded route. openpilot is unranked: it often stalls (Rc 0.217), which inflates NC and TTC.
\end{minipage}
\end{table*}

\subsection{Setup}
\label{sec:setup}
\textbf{Benchmarks.}
We evaluate in closed loop on AlpaSim~\cite{alpasim_challenge}, HUGSIM~\cite{hugsim}, and Odyssey~\cite{odyssey}, which render reconstructed real scenes along the trajectory of the policy, and in open loop on NAVSIM~\cite{navsim}.
Closed-loop rollouts are collected in AlpaSim only, and all AlpaSim results are on the PAI-AV track of the challenge: 10\,Hz, with the other agents replaying the log.
A rollout scores zero on an at-fault collision, on leaving the road, or on a lateral exit of 4\,m or more from the route corridor, and otherwise $\min(p/0.8,\,1)$, where $p$ is its progress relative to the log.
The local exam has 433 scenes that are disjoint from the 719 training scenes, with one rollout per scene; we report the mean score, the zero-score scenes (zeros), and the zeros caused by an at-fault collision or by leaving the road (at-fault).
The exam is our development set: the recipe, the serving rules, and the submitted seed were selected on it.
The held-out test on AlpaSim is the private leaderboard of the challenge, which ranks entries by the Policy Capability Score (PCS), an item-response-theory estimate of policy ability~\cite{driveirt}.
The other three benchmarks were not used for training or selection: HUGSIM (436 scenarios), the non-reactive Odyssey board (100 routes), and NAVSIM navtest; the overlaps of the training logs with Odyssey and NAVSIM are stated in Sec.~\ref{sec:main}.

\begin{figure*}[t]
\centering
\includegraphics[width=\textwidth]{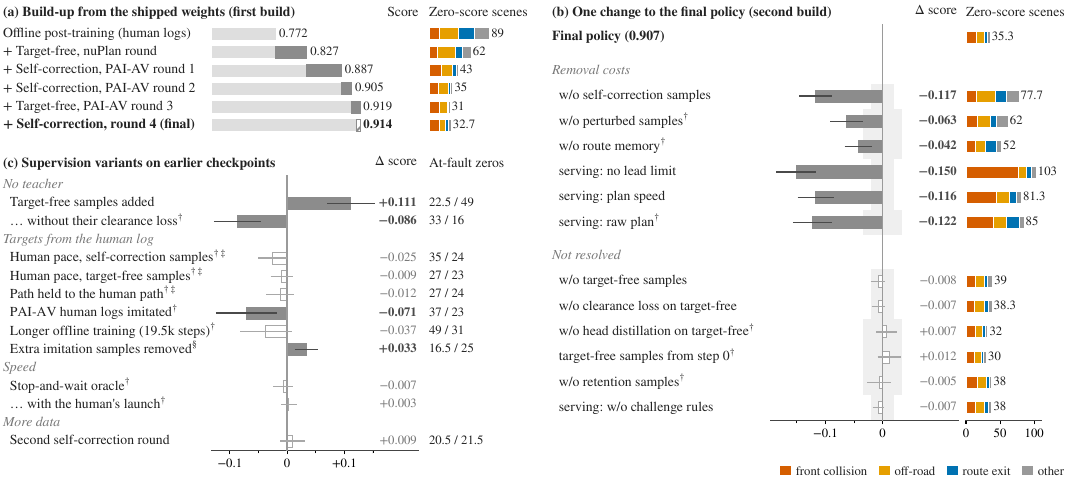}
\caption{Ablation on the AlpaSim exam (433 scenes in (a, b)). (a) Sequential build-up from the shipped weights (first build; one read per row; final: three seeds); dark grey shows the gain over the row above and hatched shows a decrease. (b) Single changes to the final training or serving layer (second build), paired with the same-seed final policy and grouped by outcome; shaded bands mark the comparison resolution (0.020 for three seeds, 0.034 for one), and filled bars exceed it. (c) Controlled pairs on earlier checkpoints (two seeds unless marked) with at-fault zeros (arm / control); filled bars have 95\,\% CIs excluding zero. Right bars in (a, b) show mean zero-score scenes per seed by cause. $^{\dagger}$One seed. $^{\ddagger}$431-scene exam (other rows in (c) use its 281-scene subset). $^{\S}$Original challenge scorer.}
\label{fig:ablation}
\end{figure*}

\textbf{Statistics.}
Differences are paired per scene with 95\,\% bootstrap intervals; on the exam, a difference counts as resolved beyond 0.020 with three training seeds and beyond 0.034 with one, estimated from 34 reads of development checkpoints (standard deviation over training seeds 0.009).
The recipe was built twice from the shipped weights, with three final seeds each, and the two builds agree ($+0.007$ $[-0.004, +0.017]$).
Fig.~\ref{fig:ablation}(a) reads the first build, the only one with a read of every stage (final policy 0.914, 32.7 zeros); every other table and figure reads the second (0.907, 35.3 zeros; seeds 0.908, 0.912, 0.902), whose rollouts the arms of Fig.~\ref{fig:ablation}(b) reuse.

\textbf{Baselines.}
We compare against a representative open-source planner from each major family of end-to-end driving: UniAD~\cite{uniad} (modular, bird's-eye view), AutoVLA~\cite{autovla} (vision-language-action), DrivoR~\cite{drivor} (trajectory scoring), WA-JEPA~\cite{wa_jepa} (world model), and py123d\_garage~\cite{garage}, the TransFuser model that the challenge organizers post-train in AlpaSim.
All run unchanged with their released weights; each is served the camera rig, frame timing, and ego state of its training data, resampled from the cameras of the benchmark.
Our harness reproduces the published per-scene results of py123d\_garage to within $+0.014$ $[-0.009, +0.037]$.
The shipped openpilot model runs through the serving stack of Lao Siji with its weights unchanged; as it has no route input, the navigation command of each closed-loop benchmark is fed as its turn desire.

\textbf{Compute.}
All GPU-hours are measured on NVIDIA RTX 6000D cards.
Collection and training take about 36 GPU-hours for the whole chain (34 for a single final policy), evaluation excluded; the final training takes about one GPU-hour with a 41\,GB memory peak.
The policy runs on a single NVIDIA RTX 3090 with a median latency of 42\,ms per frame and 1.3\,GiB of GPU memory.

\subsection{Comparison with Baselines}
\label{sec:main}
\textbf{AlpaSim.}
The checkpoint submitted to the challenge was trained with this recipe during its development and ranks third on the private leaderboard (PCS 2497).
The recipe as released, rebuilt from the shipped weights, is level with the submitted one on the local exam (a difference of $-0.004$, not resolved) and scores 0.907 over three seeds (Table~\ref{tab:main}).
The rows separate by training in the simulator.
The four planners that have not seen it, one from each family, fail at least half of the scenes, so their rows measure the domain gap as much as the planner.
The shipped openpilot model fails more than half as well (0.377), and post-training lifts it to 0.907 through the same serving stack.
Of the two models post-trained in these scenes, Lao Siji is $+0.133$ $[+0.095, +0.172]$ above py123d\_garage while training only 5\,\% of its own parameters, and has half as many at-fault zeros.
Closed-loop training also changes the kind of failure (Table~\ref{tab:cause}): lateral zeros outnumber front collisions by 2.5 to 3.8 times in every model without it, and the two models with it bring them level (py123d\_garage 29 against 31, Lao Siji 16 against 15).

\begin{table}[t]
\centering
\caption{Policy versus serving layer on the 433-scene AlpaSim exam.}
\label{tab:serving}
\footnotesize
\setlength{\tabcolsep}{2.5pt}
\resizebox{\linewidth}{!}{\begin{tabular}{@{}lccc@{}}
\toprule
(a) Policy & Raw plan & Full layer & Full $-$ raw \\
\midrule
openpilot & 0.245 & 0.377 & $+0.132$\,{\scriptsize$[+0.089, +0.176]$} \\
\quad after offline post-training & 0.466 & 0.766 & $+0.300$\,{\scriptsize$[+0.256, +0.346]$} \\
Lao Siji (final release) & 0.785 & 0.908 & $+0.122$\,{\scriptsize$[+0.088, +0.157]$} \\
\midrule
(b) Serving & Lao Siji & py123d\_garage & Difference \\
\midrule
Raw plan & 0.785 & 0.774$^{\circ}$ & $+0.011$\,{\scriptsize$[-0.035, +0.058]$} \\
Plan speed & 0.790 & 0.770 & $+0.020$\,{\scriptsize$[-0.027, +0.067]$} \\
No lead limit & 0.763 & 0.756 & $+0.006$\,{\scriptsize$[-0.036, +0.050]$} \\
Full layer & 0.908 & n/a & -- \\
\bottomrule
\end{tabular}
}\\[2pt]
\begin{minipage}{\linewidth}
\scriptsize One read per cell (Lao Siji: seed 0); differences are paired with 95\,\% intervals. Raw plan: plan passed directly to the simulator controller; Plan speed: our layer without challenge rules, re-timed to the plan's speed profile; No lead limit: without challenge rules and $v_{\mathrm{lead}}$ (inapplicable to py123d\_garage, which has no lead output). $^{\circ}$As served by its authors.
\end{minipage}
\end{table}

The lead over py123d\_garage is a lead of the served system (Table~\ref{tab:serving}).
With its raw plan handed to the controller, Lao Siji is level with py123d\_garage as its authors serve it, and the two remain level under our layer without the lead limit (b); the layer itself neither helps nor hurts py123d\_garage (0.756 under our layer and 0.770 with the plan speed, against 0.774 under its own serving; neither difference is resolved).
The lead thus comes with the lead limit, which needs a lead estimate: openpilot ships one and post-training preserves it, whereas py123d\_garage has no such output.
The gain of closed-loop post-training does not depend on the layer (a): with the raw plan it is $+0.319$ $[+0.271, +0.369]$ over offline post-training, more than under the full layer ($+0.142$).

\textbf{Zero-shot transfer (HUGSIM, Odyssey).}
The gain of post-training carries over to benchmarks on which nothing was trained.
On HUGSIM it lifts the shipped model from 0.182 to 0.460 (a paired difference of $+0.280$ $[+0.243, +0.317]$ on the 433 scenarios scored in every read), and the three leading models are not separated there (WA-JEPA 0.461, Lao Siji 0.460, py123d\_garage 0.457; the interval of one read is about $\pm 0.03$).
The lift comes with route completion (Rc 0.217 to 0.535): the shipped model stalls short of the goal, and the post-trained policy drives the route.
What is still missing shows in the sub-scores: the collision scores (NC, TTC) are about 0.02 below the best model, the drivable-area compliance 0.035.
On Odyssey, post-training lifts the shipped model from 11.2 to 49.4 RouteDS, behind only py123d\_garage, which receives target points from the recorded route.
Odyssey is not a clean zero-shot test for our policy: the nuPlan rollout round shares logs with 68 of its 100 routes.
Relative to the baselines, the policy shows no advantage on these routes ($+1.4$ RouteDS), but with an interval of about $\pm 13$ an advantage cannot be excluded.

\textbf{Open loop (NAVSIM).}
The open-loop order differs from the closed-loop one.
On NAVSIM, where the navigation command is the only route input, post-training raises the EPDMS of the shipped model from 80.5 to 88.7, which leaves Lao Siji and py123d\_garage behind AutoVLA, DrivoR, and WA-JEPA.
On AlpaSim and Odyssey the two models post-trained in closed loop lead, and DrivoR, first in PDMS, is no better than third on the headline score of any closed-loop benchmark.
With six models, two of them trained in the simulator of one benchmark, we report this as an observation and compute no rank statistic; NAVSIM is also in domain for the offline part of our recipe, whose human samples are its navtrain split.

\subsection{Ablation Study}
\label{sec:ablation}\label{sec:chain}\label{sec:carry}\label{sec:speed}\label{sec:nothelp}\label{sec:failure}
\begin{figure}[t]
\centering
\includegraphics[width=\linewidth]{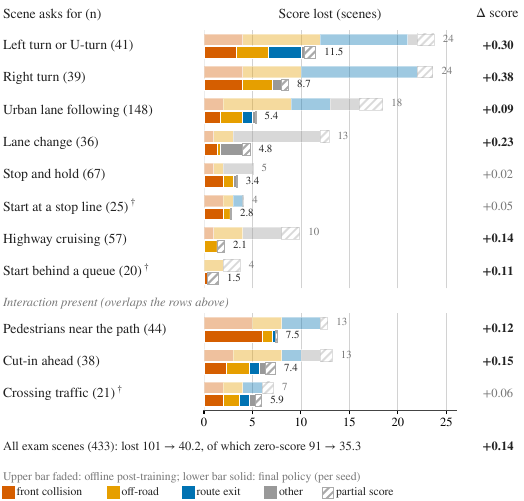}
\caption{Capability profile on the 433-scene exam. Bars sum $1-\text{score}$ within each stratum for offline post-training and the final policy (three-seed mean). Right: gain in mean score ($\Delta$, bold when its 95\,\% paired CI excludes zero). $^{\dagger}n < 30$.}
\label{fig:capability}
\end{figure}

This section asks which parts of the recipe contribute to the score, which do not, and in which scenes the gain appears.
Fig.~\ref{fig:ablation} holds the ablations: panel (a) follows the training chain round by round (Sec.~\ref{sec:rollouts}), panel (b) changes one setting of the final training or of the serving layer and pairs each arm with the final policy of the same seeds, and panel (c) collects controlled pairs on earlier checkpoints for variants that the final recipe does not contain.
The arms of (b) reuse the rollouts of the full recipe, so they measure the contribution of a component to the final training only.
Fig.~\ref{fig:capability} splits the exam by what a scene asks of the driver.
A difference within the resolution of its comparison (Sec.~\ref{sec:setup}) is reported as not resolved; in panel (c), whose pairs use other checkpoints and exam subsets, a difference counts as resolved when its 95\,\% interval excludes zero.

\textbf{Effective components.}
Six changes in panel (b) cost more than the resolution: three in the training samples and inputs, three in the serving layer.
The five closed-loop rounds add $+0.142$ $[+0.107, +0.176]$ to offline post-training (a), most of it in the first two rounds; the three later rounds add $+0.027$ $[+0.006, +0.049]$ together, and the last round alone is not resolved ($-0.006$).
The self-correction samples carry this gain: without them the final training loses $-0.117$ $[-0.145, -0.089]$, and its zero-score scenes return to the profile of offline post-training.
Removing the perturbed samples costs $-0.063$ $[-0.091, -0.034]$ although every rollout remains.
A possible reason is coverage, which we did not test: perturbations surround every logged frame, whereas rollouts hold only the states that an earlier policy reached.
Emptying the route memory at inference costs $-0.042$ $[-0.066, -0.019]$, mainly through route exits (5 to 16 zeros).

The largest effect is in the serving layer.
Its three arms also drop the two challenge rules, which alone cost $-0.007$.
Without the lead limit, the at-fault front collisions rise from 15 to 76 scenes and the score falls by $-0.150$ $[-0.185, -0.116]$.
Serving the speed profile of the plan costs $-0.116$ $[-0.149, -0.085]$, and handing the raw plan to the controller costs $-0.122$ $[-0.157, -0.088]$ on one seed (Table~\ref{tab:serving}).
These two arms are level in score, and the raw plan adds a loss of control: the car spins in 32 scenes, against 4 with the plan speed and none with the full layer.
No training arm moves the front collisions, which stay between 12 and 18 scenes in every row of (a) and in every training arm of (b).
Longitudinal safety therefore rests on the lead estimate of the shipped model and on the rule that uses it, and closed-loop post-training improves the lateral side.
The recipe gives the policy no deceleration target, and whether braking can be learned in closed loop was not tested.
The three serving arms also move the policy off the speeds at which its rollouts were collected.

\textbf{Components without a resolved effect.}
Six changes in panel (b) stay within the resolution.
Removing the target-free samples from the final training changes the score by $-0.008$, and switching off their clearance loss by $-0.007$.
On a policy trained without teacher targets, the same samples are worth $+0.111$ $[+0.070, +0.154]$, and switching their loss off takes most of this back (c): states reached by the policy help only when they are supervised.
The clearance loss thus replaces the teacher where no target trajectory exists, and with teacher targets in the recipe its contribution falls within the resolution; the self-correction samples carry the same clearance penalties, which were not ablated.
The two kinds also differ in the number of samples (273k against 41k) and in their share of a batch (9\,\% against 25\,\%), so these arms do not separate the kind of target from the amount of data.
The head distillation on target-free samples, the entry step of these samples, the retention samples, and the two challenge-specific serving rules are within the resolution as well ($+0.007$, $+0.012$, $-0.005$, $-0.007$).
We keep these components as released and make no claim for them; whether the retention samples keep the network without the adapter equal to the shipped model, which is their purpose, was not measured.

\textbf{Variants that did not help.}
Panel (c) also holds variants that were tried during development and left out of the recipe, most of them on one seed.
Six of them take more or less supervision from the human log.
Two are resolved: imitating the human logs of the training scenes directly costs $-0.071$ $[-0.124, -0.017]$, and removing a block of additional imitation samples gains $+0.033$ $[+0.014, +0.054]$.
The other four are not resolved in score (the human pace as the speed target on either closed-loop kind, the human path as the lateral target, and a longer offline imitation), and each adds 3 to 18 at-fault zeros.
The longer offline training has the best open-loop displacement error of its family and adds the most at-fault zeros.
On a state that the human never visited, the log describes another position, another time, and another phase of the replayed traffic; the recipe takes one quantity from it, the lateral placement within the lane.
A second self-correction round on top of the first is not resolved.
A speed oracle did not help either: stopping and waiting exactly where and as long as the human did is not resolved, since the scorer reads no traffic lights and the replayed agents do not react~\cite{simnet}, so that a stop on time followed by a late launch is hit by a replayed follower.

\textbf{Capability by scenario.}
Fig.~\ref{fig:capability} compares offline post-training with the final policy within each type of scene; the tags come from the human log and the lane map alone.
The gain is largest in turns ($+0.30$ for left turns, $+0.38$ for right turns), where the route exits fall from 21 to 3 scenes and the off-road ends from 14 to 6.
Lane changes gain $+0.23$, and urban lane following and highway cruising reach 0.96.
Turns nevertheless remain the weakest scenes: left turns are the lowest of the eight strata (0.72), and the 80 turn scenes hold half of the score that the final policy loses.
The same profile appears zero-shot on HUGSIM: on the 215 scenarios whose route has neither a junction turn nor a bend, the policy scores 0.504 against 0.484 and 0.420 for the other two leading models, and at turns its drivable-area compliance is 0.84 against 0.88 to 0.95.
Scenes with crossing traffic ($+0.06$) and starts at a stop line ($+0.05$) show no resolved gain, and the at-fault front collisions of the whole exam stay at 15 scenes per seed.
Over three seeds the final policy has 44 at-fault front collisions: it hits crossing or oncoming traffic in left turns (9 of 10), crossing traffic at stops and at starts from a stop line (8 of 12), and standing vehicles in right turns (7 of 12).
In 31 of the 35 collisions for which the human timing is known, the policy is ahead of the human, by 2.0\,s in the median; the pace rule of the teacher and the speed floor of the serving layer may contribute to this lead, and no arm removes them alone.

\section{Limitations and Conclusion}
\label{sec:conclusion}

\textbf{Limitations.}
The recipe supervises the lateral path only, so it teaches the policy neither to interact with other traffic nor to brake hard in edge cases.
From offline post-training to the final policy, the at-fault front collisions stay at 15 scenes, those in scenes with pedestrians near the path at 5 to 6, and the at-fault side collisions at 3; longitudinal safety rests on the lead estimate of the shipped model and a rule (Sec.~\ref{sec:speed}).
The exam is a development set, several ablations rest on a single seed, and the encoder was never unfrozen in closed loop (open-loop pilots on NAVSIM gained at most 0.17 EPDMS).

\textbf{Conclusion.}
A production driving model can be adapted to closed-loop route following by post-training 5\,\% of its parameters in reconstructed scenes, with about 34 GPU-hours of rollouts and training and a final training of one GPU-hour.
The gain comes from lateral corrective rollouts; longer training and human-log targets add nothing that the exam resolves, and braking remains with the shipped model.
Future work will analyze the remaining failure cases and explore whether post-training can also add specific abilities, such as yielding to crossing traffic and emergency braking.

\bibliographystyle{IEEEtran}
\bibliography{references}

\begin{thebibliography}{10}
\providecommand{\url}[1]{#1}
\csname url@samestyle\endcsname
\providecommand{\newblock}{\relax}
\providecommand{\bibinfo}[2]{#2}
\providecommand{\BIBentrySTDinterwordspacing}{\spaceskip=0pt\relax}
\providecommand{\BIBentryALTinterwordstretchfactor}{4}
\providecommand{\BIBentryALTinterwordspacing}{\spaceskip=\fontdimen2\font plus
\BIBentryALTinterwordstretchfactor\fontdimen3\font minus
  \fontdimen4\font\relax}
\providecommand{\BIBforeignlanguage}[2]{{%
\expandafter\ifx\csname l@#1\endcsname\relax
\typeout{** WARNING: IEEEtran.bst: No hyphenation pattern has been}%
\typeout{** loaded for the language `#1'. Using the pattern for}%
\typeout{** the default language instead.}%
\else
\language=\csname l@#1\endcsname
\fi
#2}}
\providecommand{\BIBdecl}{\relax}
\BIBdecl

\bibitem{uniad}
Y.~Hu, J.~Yang, L.~Chen, K.~Li, C.~Sima, X.~Zhu, S.~Chai, S.~Du, T.~Lin,
  W.~Wang, L.~Lu, X.~Jia, Q.~Liu, J.~Dai, Y.~Qiao, and H.~Li,
  ``Planning-oriented autonomous driving,'' in \emph{Proc. IEEE/CVF Conf.
  Computer Vision and Pattern Recognition (CVPR)}, 2023.

\bibitem{hydramdp}
Z.~Li, K.~Li, S.~Wang, S.~Lan, Z.~Yu, Y.~Ji, Z.~Li, Z.~Zhu, J.~Kautz, Z.~Wu,
  Y.-G. Jiang, and J.~M. Alvarez, ``{Hydra-MDP}: End-to-end multimodal planning
  with multi-target hydra-distillation,'' \emph{arXiv preprint
  arXiv:2406.06978}, 2024.

\bibitem{drivor}
E.~Kirby, A.~Boulch, Y.~Xu, Y.~Yin, G.~Puy, {\'E}.~Zablocki, A.~Bursuc,
  S.~Gidaris, R.~Marlet, F.~Bartoccioni, A.-Q. Cao, N.~Samet, T.-H. Vu, and
  M.~Cord, ``Driving on registers,'' \emph{arXiv preprint arXiv:2601.05083},
  2026.

\bibitem{autovla}
Z.~Zhou, T.~Cai, S.~Z. Zhao, Y.~Zhang, Z.~Huang, B.~Zhou, and J.~Ma,
  ``{AutoVLA}: A vision-language-action model for end-to-end autonomous driving
  with adaptive reasoning and reinforcement fine-tuning,'' in \emph{Advances in
  Neural Information Processing Systems (NeurIPS)}, 2025.

\bibitem{alpamayo_r1}
Y.~Wang, W.~Luo, J.~Bai, Y.~Cao, T.~Che, K.~Chen, Y.~Chen \emph{et~al.},
  ``{Alpamayo-R1}: Bridging reasoning and action prediction for generalizable
  autonomous driving in the long tail,'' \emph{arXiv preprint
  arXiv:2511.00088}, 2025.

\bibitem{epona}
K.~Zhang, Z.~Tang, X.~Hu, X.~Pan, X.~Guo, Y.~Liu, J.~Huang, L.~Yuan, Q.~Zhang
  \emph{et~al.}, ``{Epona}: Autoregressive diffusion world model for autonomous
  driving,'' in \emph{Proc. IEEE/CVF Int. Conf. Computer Vision (ICCV)}, 2025.

\bibitem{wa_jepa}
X.~Wang, Y.~Xiang, Y.~Zhou, J.~Wang, M.~Huang, J.~Huang, D.~Wei \emph{et~al.},
  ``{WA-JEPA}: Rethinking the video {JEPA} paradigm for world-action modeling
  in autonomous driving,'' \emph{arXiv preprint arXiv:2608.20974}, 2026.

\bibitem{vavam}
F.~Bartoccioni, E.~Ramzi, V.~Besnier, S.~Venkataramanan, T.-H. Vu, Y.~Xu,
  L.~Chambon \emph{et~al.}, ``{VaViM} and {VaVAM}: Autonomous driving through
  video generative modeling,'' \emph{arXiv preprint arXiv:2502.15672}, 2025.

\bibitem{codevilla2019}
F.~Codevilla, E.~Santana, A.~M. L{\'o}pez, and A.~Gaidon, ``Exploring the
  limitations of behavior cloning for autonomous driving,'' in \emph{ICCV},
  2019.

\bibitem{hiddenbiases_data}
J.~Zimmerlin, J.~Bei{\ss}wenger, B.~Jaeger, A.~Geiger, and K.~Chitta, ``Hidden
  biases of end-to-end driving datasets,'' \emph{arXiv preprint
  arXiv:2412.09602}, 2024.

\bibitem{lead}
L.~Nguyen, M.~Fauth, B.~Jaeger, D.~Dauner, M.~Igl, A.~Geiger, and K.~Chitta,
  ``{LEAD}: Minimizing learner--expert asymmetry in end-to-end driving,'' in
  \emph{CVPR}, 2026.

\bibitem{navsim}
D.~Dauner, M.~Hallgarten, T.~Li, X.~Weng, Z.~Huang, Z.~Yang, H.~Li,
  I.~Gilitschenski, B.~Ivanovic, M.~Pavone, A.~Geiger, and K.~Chitta,
  ``{NAVSIM}: Data-driven non-reactive autonomous vehicle simulation and
  benchmarking,'' in \emph{NeurIPS Datasets and Benchmarks}, 2024.

\bibitem{hugsim}
H.~Zhou, L.~Lin, J.~Wang, Y.~Lu, D.~Bai, B.~Liu, Y.~Wang, A.~Geiger, and
  Y.~Liao, ``{HUGSIM}: A real-time, photo-realistic and closed-loop simulator
  for autonomous driving,'' \emph{arXiv preprint arXiv:2412.01718}, 2024.

\bibitem{alpasim}
{NVIDIA}, ``{AlpaSim}: A modular closed-loop simulator for autonomous
  driving,'' \url{https://github.com/NVlabs/alpasim}, 2025.

\bibitem{odyssey}
J.~Kim, H.~Shin, S.~Yu, H.~Yoo, M.~Kim, J.~Oh, D.~Kwak, S.~Nam, H.~Oh, H.~Kim,
  H.~Cho, J.~Park, S.~W. Seo, and J.~W. Choi, ``{Odyssey}: A closed-loop
  benchmark for long-horizon real-world driving with explicit navigation
  routes,'' \emph{arXiv preprint arXiv:2610.06469}, 2026.

\bibitem{alpasim_challenge}
{NVIDIA}, ``{AlpaSim} {E2E} closed-loop challenge 2026,''
  \url{https://nvidia-alpasime2eclosedloopchallenge2026.hf.space/}, 2026.

\bibitem{openpilot}
{comma.ai}, ``openpilot: An open source driver assistance system,''
  \url{https://github.com/commaai/openpilot}, 2016.

\bibitem{dagger}
S.~Ross, G.~Gordon, and D.~Bagnell, ``A reduction of imitation learning and
  structured prediction to no-regret online learning,'' in \emph{AISTATS},
  2011.

\bibitem{gigaflow}
M.~Cusumano-Towner \emph{et~al.}, ``Robust autonomy emerges from self-play,''
  \emph{arXiv preprint arXiv:2502.03349}, 2025.

\bibitem{carl}
B.~Jaeger, D.~Dauner, J.~Bei{\ss}wenger, S.~Gerstenecker, K.~Chitta, and
  A.~Geiger, ``{CaRL}: Learning scalable planning policies with simple
  rewards,'' in \emph{CoRL}, 2025.

\bibitem{rad}
H.~Gao, S.~Chen, B.~Jiang, B.~Liao, Y.~Shi, X.~Guo, Y.~Pu \emph{et~al.},
  ``{RAD}: Training an end-to-end driving policy via large-scale {3DGS}-based
  reinforcement learning,'' \emph{arXiv preprint arXiv:2502.13144}, 2025.

\bibitem{alpagym}
\BIBentryALTinterwordspacing
{NVIDIA}, ``How to post-train autonomous vehicle models in closed loop with
  {NVIDIA} {Alpamayo},'' NVIDIA Technical Blog, 2026, alpaGym. [Online].
  Available:
  \url{https://developer.nvidia.com/blog/how-to-post-train-autonomous-vehicle-models-in-closed-loop-with-nvidia-alpamayo/}
\BIBentrySTDinterwordspacing

\bibitem{lbc}
D.~Chen, B.~Zhou, V.~Koltun, and P.~Kr{\"a}henb{\"u}hl, ``Learning by
  cheating,'' in \emph{Proc. Conf. Robot Learning (CoRL)}, 2019.

\bibitem{roach}
Z.~Zhang, A.~Liniger, D.~Dai, F.~Yu, and L.~Van~Gool, ``End-to-end urban
  driving by imitating a reinforcement learning coach,'' in \emph{Proc.
  IEEE/CVF Int. Conf. Computer Vision (ICCV)}, 2021.

\bibitem{road}
G.~Garcia-Cobo, M.~Igl, P.~Karkus, Z.~Zhang, M.~Watson, Y.~Chen, B.~Ivanovic,
  and M.~Pavone, ``{RoaD}: Rollouts as demonstrations for closed-loop
  supervised fine-tuning of autonomous driving policies,'' \emph{arXiv preprint
  arXiv:2512.01993}, 2025.

\bibitem{opted}
D.~Da~Col, M.~Igl, P.~Karkus, K.~Chitta, B.~Ivanovic, M.~Pavone, K.~Schindler,
  and C.~Sakaridis, ``{OPTED}: On-policy fine-tuning for end-to-end driving
  using a render-free teacher,'' \emph{arXiv preprint arXiv:2609.20756}, 2026.

\bibitem{drivezero}
H.~He, C.~Hu, Z.~Su, H.~Zhang, H.~Liu, J.~Li, H.~Tian, Z.~Shen, H.~Li, Z.~Li,
  Y.~Yang, B.~Huang, S.~Zhang, K.~Chen, X.~Zhang, W.~Dai, H.~Dai, S.~Liu,
  Z.~Huang, and N.~Wang, ``{DriveZero}: End-to-end driving beyond human
  demonstrations,'' \emph{arXiv preprint arXiv:2609.06055}, 2026.

\bibitem{pai_av}
{NVIDIA}, ``{PhysicalAI-Autonomous-Vehicles} dataset,''
  \url{https://huggingface.co/datasets/nvidia/PhysicalAI-Autonomous-Vehicles},
  2025.

\bibitem{tfpp}
B.~Jaeger, K.~Chitta, and A.~Geiger, ``Hidden biases of end-to-end driving
  models,'' in \emph{Proc. IEEE/CVF Int. Conf. Comput. Vis. (ICCV)}, 2023, pp.
  8240--8249.

\bibitem{simlingo}
K.~Renz, L.~Chen, E.~Arani, and O.~Sinavski, ``{SimLingo}: Vision-only
  closed-loop autonomous driving with language-action alignment,'' in
  \emph{Proc. IEEE/CVF Conf. Comput. Vis. Pattern Recognit. (CVPR)}, 2025, pp.
  11\,993--12\,003.

\bibitem{garage}
{Kesai Labs}, ``py123d\_garage pretrained checkpoints,''
  \url{https://huggingface.co/kesai-labs/py123d_garage_pretrained_checkpoints},
  2026.

\bibitem{nuplan}
H.~Caesar, J.~Kabzan, K.~S. Tan, W.~K. Fong, E.~Wolff, A.~Lang, L.~Fletcher,
  O.~Beijbom, and S.~Omari, ``{nuPlan}: A closed-loop {ML}-based planning
  benchmark for autonomous vehicles,'' \emph{arXiv preprint arXiv:2106.11810},
  2021.

\bibitem{pseudosim}
W.~Cao, M.~Hallgarten, T.~Li, D.~Dauner, X.~Gu, C.~Wang, Y.~Miron, M.~Aiello,
  H.~Li, I.~Gilitschenski, B.~Ivanovic, M.~Pavone, A.~Geiger, and K.~Chitta,
  ``Pseudo-simulation for autonomous driving,'' in \emph{CoRL}, 2025.

\bibitem{driveirt}
L.~Nguyen and K.~Chitta, ``Leaderboard design: {DriveIRT},''
  \url{https://ln2697.github.io/navhard-leaderboard-2/docs/}, 2026.

\bibitem{simnet}
L.~Bergamini, Y.~Ye, O.~Scheel, L.~Chen, C.~Hu, L.~Del~Pero, B.~Osi{\'n}ski,
  H.~Grimmett, and P.~Ondruska, ``{SimNet}: Learning reactive self-driving
  simulations from real-world observations,'' \emph{arXiv preprint
  arXiv:2105.12332}, 2021.

\end{thebibliography}

\end{document}